# PRISM-AH: Knowledge-Guided Reasoning over Structured Multimodal Evidence for Ambivalence and Hesitancy Recognition

Podakanti Satyajith Chary[1], Barath Parthiban[2], Pranesh Velmurugan[2], Adeeba Khan[2], and Nagarajan Ganapathy[2]

[1] Department of Engineering Science, IIT Hyderabad, India
es25resch11002@iith.ac.in
[2] Department of Biomedical Engineering, IIT Hyderabad, India
{bm25resch11010,bm26resch01003,bm25resch11004}@iith.ac.in
gnagarajan@bme.iith.ac.in

**Abstract.** Ambivalence and hesitancy denote a complex affective state that frequently precedes the delay or abandonment of health behaviour change. Video-based state assessment is demanding, since the evidence is sparse, is distributed across facial, vocal, linguistic, and bodily channels, and varies markedly between individuals. Direct application of standard multimodal fusion to the recently introduced Behavioural Ambivalence/Hesitancy (BAH) benchmark yields low accuracy, which indicates that undifferentiated feature fusion does not capture the phenomenon. In this proposed framework, PRISM-AH (Predictive Reasoning over Interacting Streams for Multimodal Ambivalence / Hesitancy recognition), departs from feature fusion. In this work a lightweight multimodal network first converts a video into structured and timestamped evidence, namely the moments of highest conflict, the dominant modality at each moment, and the associated prosodic descriptors. A large language model then reasons over this evidence under an explicit prompt that encodes the expert cue taxonomy of the benchmark, and its verdict is combined with the network prediction through calibrated late fusion. PRISM-AH attains a macro F1 of 0.6324 on the held-out private test partition, which corresponds to 2.24 times the reported zero-shot baseline of 0.2827. A controlled study over three seeds isolates the reliable sources of performance, quantifies each modality, and characterises the behaviour of the approach under missing modalities and across participant subgroups. PRISM-AH source code is publicly available in github at - https://github.com/Satyajithchary/PRISM-AH-Predictive-Reasoning-over-Interacting-Streams-for-Multimodal-A-H-recognition .

## 1 Introduction

Ambivalence and hesitancy (A/H) constitute a conflicted state that sits between positive and negative orientation toward an action, or between acceptance and refusal of it. The state is a principal reason that individuals delay or abandon changes to health behaviour [3]. An assistant that perceives hesitation during a digital behaviour change intervention is able to adapt its guidance, which motivates the automatic recognition of A/H from video.

Expert annotators identify A/H through cues that are distributed across facial, linguistic, vocal, and bodily channels [3]. The most reliable evidence arises when two channels disagree at the same instant, for example an affirmation in speech accompanied by an averted gaze or a shaking head. Conflict confined to a single channel also signals A/H, and its detection is more demanding. The Behavioural Ambivalence/Hesitancy (BAH) benchmark [3] was introduced to study the phenomenon, and the accompanying analysis established that standard multimodal fusion reaches only a low level of accuracy on the task. That observation frames the central difficulty. The discriminative evidence is sparse and localised in time, individuals differ in how hesitation surfaces, and the annotated corpus is modest, which together penalise pipelines that fuse dense per-frame features without regard to structure.

The proposed work addresses this difficulty through a change of representation. Rather than pool raw multimodal features into a single vector, PRISM-AH - Predictive Reasoning over Interacting Streams for Multimodal Ambivalence / Hesitancy recognition extracts a compact and interpretable summary of a video, namely the moments at which cross-channel conflict is strongest, the channel that dominates each moment, the associated prosodic and disfluency descriptors, and the verbatim transcript. A large language model (LLM) then reasons over this structured evidence under a prompt that encodes the expert cue taxonomy of the benchmark. The reasoning verdict is combined with the discriminative prediction through late fusion, and the fusion is retained only when it improves validation performance. The design places domain knowledge at the point of decision, where a flexible reasoner is able to weigh conflicting cues in the manner of an expert, and it produces a rationale that cites the decisive moments.

The contribution of the proposed work is three fold -

(i) A recognition approach that reformulates video-level A/H detection as reasoning over structured, timestamped multimodal evidence rather than as fusion of dense features, and it attains a macro F1 (mF1) of 0.6324 on the private test partition of the BAH benchmark, corresponding to 2.24× the reported zero-shot baseline of 0.2827.

(ii) A knowledge-guided reasoning layer, in which an LLM interprets the emitted evidence under an explicit encoding of the expert cue taxonomy. A controlled study establishes that the encoded taxonomy is necessary for the improvement, since its removal from the prompt eliminates the gain, and that a reasoner of sufficient capacity is required.

(iii) A rigorous empirical analysis over three experimental seeds that quantifies modality contributions and evaluates robustness across participant subgroups and missing channels, establishing language as the primary driver of predictive performance.

## 2 Related Work

**Recognition of ambivalence and hesitancy.** The BAH benchmark [3] introduced video-level and frame-level recognition of A/H, together with expert cue labels, timestamped transcripts, cropped faces, and participant metadata. The reported baseline applies a zero-shot multimodal LLM, Video-LLaVA [6], to the vision channel under a simple prompt, and reaches a macro F1 of 0.2827. The strength of that baseline lies in its generality, since it requires no training. Its limitation is equally clear, because it ignores the audio and language channels, applies no domain knowledge, and produces no localisation of the cues that drive the decision. A second reference model fuses vision, audio, and text for frame-level prediction, and the benchmark analysis reports that such fusion attains a low level of accuracy at the video level. The proposed work retains the multimodal signal of the second model and the flexibility of a reasoner, and it addresses the missing element, namely the explicit use of the expert cue structure at the point of decision [3].

**Multimodal representation and temporal modelling.** Frozen foundation encoders provide strong per-channel representations at low cost, and attention supports coupling across channels and across time [2, 8, 11, 13]. The strength of these representations is their transferability. Their limitation on the present task is that a pooled multimodal vector discards the temporal locality and the cross-channel disagreement that define A/H, which is consistent with the low accuracy reported for standard fusion [3]. The proposed work therefore preserves the per-window structure and converts it into an explicit conflict signal that a reasoner is able to interpret.

**Reasoning with large language models.** Instruction-tuned LLMs reason effectively over structured textual inputs, and efficient serving makes batched inference practical [5, 12]. Prior application to affective video, as in the benchmark baseline [6], prompts a model with raw or lightly processed input and reports weak accuracy on A/H. The limitation of that usage is the absence of domain grounding and of a compact evidence representation, which leaves the model to infer subtle cues from unstructured input. The proposed work differs in two respects: (i) the reasoner receives a distilled and timestamped evidence summary rather than raw frames and, (ii) the prompt encodes the expert cue taxonomy, which supplies the domain knowledge that the benchmark analysis identifies as necessary.

**Adaptation and personalisation.** Since the benchmark is partitioned by participant, adaptation to unseen individuals is pertinent, and prior work studies source-free adaptation, subject-based adaptation, and test-time cache personalisation for facial expression [9, 10, 15]. Feature-wise conditioning [7] and entropy minimisation at test time [14] provide lightweight mechanisms in this family.

**Table 1:** Data partitions. The public test partition provides labels and supports the controlled analysis. The private test partition is held out and scored by the organisers.

| Partition | Videos | Positive rate | Role |
|---|---|---|---|
| Train | 758 | 0.496 | optimisation |
| Validation | 124 | 0.605 | calibration and model selection |
| Public test (labelled) | 525 | 0.606 | controlled analysis |
| Private test | 152 | held out | leaderboard |

The empirical study reported below examines whether conditioning on participant attributes transfers to the participant-disjoint partition, and finds that it does not, which motivates a configuration that omits it.

## 3 Method

### 3.1 Problem Formulation and Benchmark

The task is a binary decision at the video level, namely whether a video contains A/H. The BAH benchmark [3] comprises 1,427 videos with a total duration of 10.60 hours from 300 participants, annotated at the frame and video level with expert cues, and accompanied by timestamped transcripts, cropped faces, and participant metadata. The benchmark is partitioned by participant, so that no individual appears in more than one partition, which places a genuine generalisation demand on any model. The partitions used in the present study are summarised in Tab. 1. The performance measure is the macro F1 across the two classes, and the average precision (AP) of the positive class is reported additionally, in accordance with the benchmark protocol [3]. The public test partition provides labels and supports the controlled analysis, whereas the private test partition is held out and scored by the organisers.

### 3.2 Overview and Rationale

The benchmark analysis indicates that A/H is not well captured by fusion of dense per-frame features [3]. Two properties explain the difficulty. The evidence is local in time, so that averaging over a clip dilutes it, and the strongest evidence is a disagreement between channels rather than a value within any single channel. The proposed approach is organised around these two properties. A lightweight network preserves temporal locality and computes an explicit measure of cross-channel disagreement, and it emits a compact evidence summary. A reasoner then applies domain knowledge to that summary. Figure 1 presents the complete approach.

### 3.3 Structured Evidence Extraction

A video is divided into windows of 0.5 seconds, which matches the average A/H segment length reported for the benchmark and preserves the temporal locality of the evidence. For each window, aligned face frames are encoded with a vision transformer (ViT) trained under language supervision [8], speech is encoded with a self-supervised speech model WavLM [2], and the transcript segment is encoded with a sentence encoder MPNet [11]. Prosodic and disfluency descriptors, namely pitch variation, energy, pause structure, filled pauses, hedges, and contrastive markers, are computed per window in accordance with the language and audio cue definitions of the codebook [3]. The encoders remain frozen, which suits the modest size of the corpus, and the per-video features are cached once.

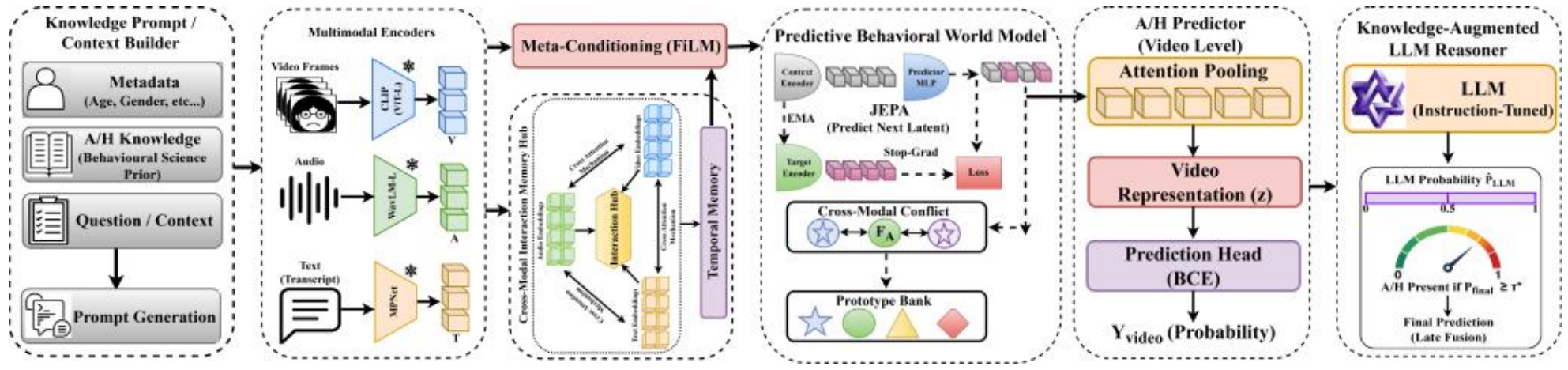

**Fig. 1:** Overview of PRISM-AH. Frozen encoders produce time-aligned window features that are conditioned on participant metadata through feature-wise modulation. A cross-modal interaction hub with temporal memory couples the channels, a predictive component exposes a hesitation surprise signal, an explicit conflict measure scores cross-channel disagreement, and a prototype bank matches recurring patterns. An attention-pooled head produces the calibrated video-level probability, and the same evidence is distilled into a textual summary. A knowledge-guided language model reasons over the summary and contributes a second opinion through late fusion.

Each channel is projected to a shared dimension and modulated by participant metadata through feature-wise linear modulation (FiLM) [7]. A stack of interaction blocks then couples the three channels with a shared hub state and a temporal memory. Each block applies cross-attention between the hub and the channels, followed by temporal self-attention, so that information propagates across channels and across time. Cross-channel disagreement is measured explicitly. For each window the projected channel vectors are compared through pairwise cosine dissonance,

$$d_t^{va} = 1 - \cos(\mathbf{v_t}', \mathbf{a_t}'),\ d_t^{v\tau} = 1 - \cos(\mathbf{v_t}', \boldsymbol{\tau}_\mathbf{t}'),\ d_t^{a\tau} = 1 - \cos(\mathbf{a_t}', \boldsymbol{\tau}_\mathbf{t}'), \quad (1)$$

where $\mathbf{v_t}'$, $\mathbf{a_t}'$, and $\boldsymbol{\tau}_\mathbf{t}'$ denote the projected vision, audio, and text features. The dissonance triplet drives a conflict head that produces a per-window conflict score, which realises the codebook definition of cross-modal inconsistency as strong evidence of A/H [3].

Two further signals enrich the evidence. A predictor anticipates the next hub state, and an exponential moving average of the hub supplies a stable target, in the spirit of a joint-embedding predictive architecture (JEPA) [1]. The prediction error at each window,

$$s_t = \|\hat{\mathbf{h}}_{t+1} - \bar{\mathbf{h}}_{t+1}\|_2^2, \quad (2)$$

serves as a hesitation surprise signal, since a break in behavioural continuity is characteristic of A/H. A bank of learnable prototypes captures recurring hesi-tation patterns, and each window is assigned to prototypes by cosine similarity, which yields an interpretable pattern label for the emitted evidence.

### 3.4 Discriminative Head and Objective

An attention-pooled head aggregates the window representations into a video representation, and a linear layer produces the video logit, which is the ranked prediction. The window-level annotations are exploited through an auxiliary head that predicts A/H presence per window, which supplies a dense training signal that a video-only objective lacks. The complete objective combines the video term with the auxiliary terms,

$$L = w_v L_{\text{video}} + w_f L_{\text{frame}} + w_j L_{\text{jepa}} + w_c L_{\text{conflict}} + w_p L_{\text{proto}}, \tag{3}$$

with weights $w_v$=1.0, $w_f$ =0.5, $w_j$=0.3, $w_c$=0.1, and $w_p$=0.05. The video and window terms are binary cross-entropy with label smoothing.

### 3.5 Calibration and Test-Time Adaptation

Since the metric is macro F1, the decision threshold is calibrated on the validation partition to maximise macro F1, which corrects for the tendency of the model to over-predict the positive class. Test-time adaptation (TTA) is available as an option, in which a small number of entropy-minimisation steps update the affine parameters of the normalisation layers [14] without labels.

### 3.6 Knowledge-Guided Reasoning and Late Fusion

The discriminative network distils each video into a compact evidence summary, namely the discriminative probability, the number of windows, the matched prototype, the windows of highest A/H probability with their timestamps and dominant modality and prosodic descriptors, and the transcript. An instruction-tuned LLM [12] receives this summary under the prompt shown in Fig. 2, which encodes the expert cue taxonomy of the codebook, and returns a verdict with an associated confidence. The reasoner operates in a zero-shot manner and is served through efficient batched inference [5]. The design supplies the reasoner with the two elements that the benchmark baseline lacks, namely a compact evidence representation and explicit domain knowledge.

The reasoning probability $q$ is combined with the discriminative probability $p$ through convex late fusion,

$$p_{\text{final}} = (1 - \omega)\, p + \omega\, q, \tag{4}$$

where the weight $\omega$ and the fused threshold are selected on the validation partition. Fusion is adopted only when it improves validation macro F1, which guards against a reasoner that would otherwise degrade a strong discriminative prediction. Figure 3 illustrates the evidence-to-verdict format.

**Knowledge-guided reasoning prompt (condensed)**

[A/H codebook] Ambivalence/Hesitancy is the simultaneous presence of competing positive and negative feelings, expressed as CONFLICT across or within modalities, judged against the person's own baseline.
FACIAL: gaze shifts, frown, brow raise/lower, lip-press, pout, fake smile, blink-rate change.
LANGUAGE: filler sounds (umm, uh); filler words (like, you know); hedging (I think, not sure); self-correction; contrastive 'but'.
AUDIO: silent pauses, cut-off words, slowing, shaky voice, sigh, tsk-click, stutter.
BODY: look-away, head-shake, tilt, shrug, self-touch, restless motion.
CROSS-MODAL (strongest): a modality contradicting another at the same moment.

[Participant] age=?, gender=?, question #k of a fixed set.

[Model evidence] discriminative probability; number of 0.5s windows; matched prototype; highest-A/H moments as {start-end s / prob / dominant modality / prosody}.

[Transcript] verbatim words for the clip.

[Instruction] Reason over the multimodal conflict and temporal pattern, then output exactly one line:
VERDICT: {"ah": 0/1, "confidence": 0.0-1.0, "reason": "<=25 words citing the decisive cues and timestamps"}

**Fig. 2:** The knowledge-guided reasoning prompt. The reasoner receives the expert cue taxonomy, the participant context, and the structured timestamped evidence, and returns a single verdict.

## 4 Experiments

### 4.1 Implementation Details

Aligned faces are encoded with a ViT of the CLIP family [8], speech with a large self-supervised speech model wavlm [2], and transcripts with a sentence encoder mpnet [11]. Window features are cached to disk. The network is trained with a learning rate of $2 \times 10^{-4}$, weight decay of $10^{-2}$, and gradient clipping. Optimisation is conducted in single precision, which avoids numerical instabilities observed under mixed precision on this task. The reasoner is an instruction-tuned model of the Qwen family Qwen 2.5-7B-Instruction [12], served through efficient batched inference vLLM [5] in a zero-shot configuration. The controlled analysis repeats training over three seeds and reports the mean and standard deviation. The controlled analysis is independent of the single submission run, so its absolute values differ marginally, while the relative effects are the object of study.

### 4.2 Main Results

Table 2 reports the principal outcome. The complete approach attains a macro F1 of 0.6324 on the private test partition, which more than doubles the reported zero-shot baseline of 0.2827 [3]. On the public test partition, the discriminative network reaches 0.5790, and knowledge-guided reasoning with late fusion raises the score to 0.6133. The fusion was adopted after it improved validation macro F1 from 0.6129 to 0.6301, and the improvement transferred to the public test

**Illustrative reasoning trace (format)**

```
evidence discriminative prob = 0.58; windows = 34; prototype = 4;
  1.5-2.0s / 0.81 / language / hedges=2 contrasts=1;
  4.0-4.5s / 0.74 / audio / pause, pitch drop;
  transcript: "um well maybe i think so but i am not sure"

verdict {
  "ah": 1, "confidence": 0.79,
  "reason": "hedging with contrastive ‘but’ at 1.5-2.0s and a filled pause at 4.0s
indicate unresolved stance"
}
```

**Fig. 3:** Illustration of the evidence-to-verdict format. The rationale cites the decisive cues and timestamps, which supports interpretation of the decision. The example depicts the format and does not report a specific benchmark video.

**Table 2:** Main results. Macro F1 is the primary metric. The baseline is the reported zero-shot multimodal LLM [3]. The private test value is the organiser-reported leaderboard score of the submitted approach.

| Method | Public test mF1 | Private test mF1 |
|---|---|---|
| Zero-shot baseline [3] | 0.2827 | n/a |
| PRISM-AH (discriminative) | 0.5790 | n/a |
| **PRISM-AH + knowledge-guided reasoning** | **0.6133** | **0.6324** |

partition, which indicates that the reasoning layer contributes signal beyond the small validation partition.

The complete approach attains 0.6324 macro F1 on the held-out private test partition, which corresponds to 2.24× the reported zero-shot baseline of 0.2827. Knowledge-guided reasoning improves the discriminative prediction on the public test partition from 0.5790 to 0.6133, and the improvement transfers from validation to the public test partition.

### 4.3 Analysis of the Reasoning Layer

The reasoning layer is the principal contribution, and two controlled studies establish the conditions under which it helps. Both studies operate on a fixed discriminative checkpoint, whose public-test macro F1 is 0.5825, so that the effect of the reasoner is isolated. The absolute base differs marginally from the submission run in Tab. 2, since the checkpoint is independent, while the effect of the reasoner is the object of study.

Table 3 reports a sweep over reasoning models. A model of insufficient capacity does not improve the prediction, since the fusion weight selected on validation is zero for the seven-billion-parameter model and the score is unchanged. A capable model of the same family raises the public-test macro F1 from 0.5825 to 0.6070, and a model of a different family raises it to 0.5934. The gain therefore requires a sufficiently capable reasoner.

**Table 3:** Reasoning model sweeps on the public test partition, on a fixed discriminative checkpoint whose base macro F1 is 0.5825. The fusion weight is selected on the validation partition, and a weight of zero indicates that fusion was not adopted.

| Reasoning model | Public test macro F1 | Fusion weight |
|---|---|---|
| None (discriminative only) | 0.5825 | n/a |
| Qwen2.5-7B [12] | 0.5825 | 0.0 |
| Mistral-7B [4] | 0.5934 | 0.1 |
| **Qwen2.5-14B** [12] | **0.6070** | 0.3 |

**Table 4:** Prompt ablation on the public test partition, with the strongest reasoner. Removal of the encoded expert cue taxonomy eliminates the gain, which identifies the domain knowledge as the essential ingredient.

| Prompt configuration | Public test macro F1 |
|---|---|
| Discriminative only | 0.5825 |
| Codebook knowledge and structured evidence | 0.6070 |
| Without codebook knowledge | 0.5825 |
| Without structured evidence | 0.6147 |

Table 4 reports a prompt ablation on the strongest reasoner. Removal of the expert cue taxonomy from the prompt eliminates the gain, since the selected fusion weight returns to zero and the score returns to 0.5825. Removal of the structured evidence, in contrast, retains the gain, at 0.6147. The encoded domain knowledge is therefore the essential ingredient of the reasoning layer, whereas the structured evidence is not necessary for the gain. The observation is consistent with the finding that language is the decisive channel, since the reasoner operates over the transcript together with the domain knowledge.

The gain from the reasoning layer depends on the encoded expert cue taxonomy. Its removal from the prompt eliminates the gain, from 0.6070 to 0.5825, whereas removal of the structured evidence retains the gain at 0.6147. A reasoner of sufficient capacity is required, since a seven-billion-parameter model yields no improvement.

## 4.4 Ablation Study of Structural Components

Table 5 reports a controlled leave-one-out study over three seeds on the discriminative network, evaluated at the validation-calibrated threshold. Two findings follow. Firstly, conditioning on participant attributes does not transfer to the participant-disjoint partition. Removal of the metadata conditioning raises the public-test macro F1 from 0.5570 to 0.6140, which indicates that the attributes fit the training distribution rather than the held-out one. Secondly, a compact configuration that retains the attention-pooled head over the three frozen channels reaches 0.6304, and the difference relative to the full configuration is significant under a McNemar test at $p$=0.042. The auxiliary signals, namely the conflict score, the predictive surprise, and the prototype bank, contribute marginally in this discriminative-only setting, with differences that fall within the seed variation. The finding is informative rather than adverse, since it identifies the compact configuration as the preferable discriminative backbone and concentrates the added value in the calibrated representation and in the reasoning layer.

**Table 5:** Leave-one-out study over three seeds on the public test partition. One component is removed per row. Values are mean ± standard deviation. Δ is the change in public-test macro F1 relative to the full configuration.

| Configuration | Val macro F1 | Public test macro F1 | Δ |
|---|---|---|---|
| Full configuration | 0.6096 ± 0.0171 | 0.5570 ± 0.0130 | +0.0000 |
| without dense supervision | 0.5808 ± 0.0116 | 0.5511 ± 0.0067 | −0.0059 |
| without conflict score | 0.5769 ± 0.0102 | 0.5456 ± 0.0115 | −0.0115 |
| without predictive surprise | 0.5926 ± 0.0260 | 0.5550 ± 0.0184 | −0.0021 |
| without prototypes | 0.6062 ± 0.0081 | 0.5602 ± 0.0319 | +0.0032 |
| without metadata | **0.6909 ± 0.0140** | 0.6140 ± 0.0121 | +0.0570 |
| compact head | 0.6592 ± 0.0024 | **0.6304 ± 0.0099** | +0.0734 |

**Table 6:** Modality contribution over three seeds on the public test partition. Single-modality and modality-pair models isolate each contribution. Values are mean ± standard deviation.

| Configuration | Val macro F1 | Public test macro F1 |
|---|---|---|
| Vision only | 0.5850 ± 0.0069 | 0.5570 ± 0.0232 |
| Audio only | 0.5763 ± 0.0183 | 0.5563 ± 0.0285 |
| Text only | 0.5906 ± 0.0101 | 0.5514 ± 0.0479 |
| Vision and audio | 0.5902 ± 0.0178 | 0.5477 ± 0.0257 |
| Vision and text | 0.5917 ± 0.0056 | 0.5593 ± 0.0154 |
| Audio and text | 0.5891 ± 0.0112 | 0.5491 ± 0.0430 |
| All three | **0.6096 ± 0.0171** | **0.5570 ± 0.0130** |

### 4.5 Modality Contribution

Table 6 reports the contribution of each modality over three seeds. The single-modality models reach comparable macro F1, between 0.5514 and 0.5573, and the pairwise and full combinations remain within the same range. The combination does not exceed the strongest single modality by a margin that surpasses the seed variation. The result indicates that, at the level of the discriminative decision alone, the channels carry overlapping information, and it motivates the analysis of missing modalities below.

**Table 7:** Threshold calibration and test-time adaptation on the public test partition, evaluated on the same trained model.

| Configuration | macro F1 |
|---|---|
| Fixed threshold 0.5, without TTA | **0.5768** |
| Validation-calibrated threshold | 0.5735 |
| Validation-calibrated threshold, with TTA | 0.5735 |

**Table 8:** Robustness to missing modalities on the public test partition. A single channel is zeroed only at evaluation.

| Condition | macro F1 |
|---|---|
| Full input | **0.5735** |
| Language removed | 0.4982 |
| Vision removed | 0.5553 |
| Audio removed | 0.5960 |

### 4.6 Calibration and Test-Time Adaptation

Table 7 reports the effect of the decision procedure on a trained model. Threshold calibration and the fixed threshold of one half yield comparable macro F1 on the public test partition, and TTA leaves the score unchanged. The two mechanisms therefore neither help nor harm on this partition, and the reported system retains threshold calibration for its principled selection of the operating point under the macro F1 metric.

### 4.7 Robustness to Missing Modalities

Table 8 reports the behaviour of a trained model when a single channel is zeroed at evaluation, without retraining. The removal of language causes the largest degradation, from 0.5735 to 0.4982, whereas the removal of audio does not reduce performance. The result identifies language as the decisive channel for the video-level decision, and it is in accordance with the report that text is the strongest single modality on the benchmark [3]. The finding also explains the design of the reasoning layer, which operates over the transcript and the language cues.

Language is the decisive channel. Its removal at evaluation reduces macro F1 from 0.5735 to 0.4982, whereas the removal of audio does not reduce performance. The observation is in accordance with the report that text is the strongest single modality on the benchmark [3] and it motivates a reasoning layer that operates over the transcript.

### 4.8 Subgroup Behaviour

Table 9 reports macro F1 across participant subgroups at the shared validation-calibrated threshold. The scores range from 0.5540 to 0.5826, and the gap across gender and across age band remains narrow, which indicates that performance does not concentrate in a particular subgroup.

**Table 9:** Subgroup macro F1 on the public test partition, computed at the shared validation-calibrated threshold.

| Attribute | Group | Videos | Macro F1 |
|---|---|---|---|
| Gender | Female | 252 | 0.5826 |
| Gender | Male | 266 | 0.5611 |
| Age | Under 35 | 294 | 0.5760 |
| Age | 35 to 49 | 147 | 0.5737 |
| Age | 50 plus | 84 | 0.5540 |

## 5 Discussion

The controlled study clarifies the sources of performance. The reliable contributors are the frozen multimodal representation, the calibrated attention-pooled head, and the knowledge-guided reasoning layer, which together raise the public-test macro F1 to 0.6133 and attain 0.6324 on the private test. The improvement of the complete approach is attributable to the reasoning layer, and specifically to the encoded expert cue taxonomy, since removal of the taxonomy from the prompt eliminates the gain. The auxiliary structural signals of the discriminative network contribute marginally to the discriminative decision, and their value may depend on longer optimisation or on a larger annotated corpus. A compact discriminative backbone is therefore preferable, and the added value is concentrated in the reasoning layer. The robustness analysis identifies language as the decisive channel, which indicates that further gains are most likely to arise from stronger language grounding, in agreement with the design of the reasoning layer. Conditioning on participant attributes does not transfer to the participant-disjoint partition, and a configuration that omits it is preferable.

## 6 Conclusion

The proposed work, PRISM-AH, reformulates video-level ambivalence and hesitancy recognition as knowledge-guided reasoning over structured, timestamped multimodal evidence. A lightweight network preserves temporal locality and emits a compact evidence summary, and an instruction-tuned reasoner interprets that summary under an explicit encoding of the expert cue taxonomy. The approach attains a macro F1 of 0.6324 on the private test partition of the BAH benchmark, in comparison to the reported zero-shot baseline of 0.2827. A controlled study over three seeds identify the reliable sources of performance, quantifies each modality, and identifies language as the decisive channel. The analysis motivates a compact discriminative backbone and positions stronger language grounding as the most promising direction for further improvement.